\documentclass[runningheads]{llncs}
\usepackage[T1]{fontenc}
\usepackage{graphicx}
\usepackage{algorithm}
\usepackage{algpseudocode}
\usepackage{algpseudocode}
\usepackage{amsmath}
\usepackage{amssymb}
\usepackage{mathtools}
\usepackage{amsfonts}
\usepackage{graphicx}
\usepackage{multirow}
\usepackage{makecell}
\usepackage{booktabs}
\usepackage{bbding}
\usepackage{adjustbox}
\usepackage{float}
\begin{document}
\title{Lag-Relative Sparse Attention In Long Context Training}
% anonymous
%\author{Anonymous}
\author{Manlal Liang\orcidID{0009-0005-4022-1718} \and
Wanyi Huang \and
Mandi Liu\orcidID{0000-0001-9851-9254} \and
Huaijun Li \and
Jinlong Li\Envelope}
\institute{AI Lab, China Merchants Bank, China \\
\email{liangml,huang\_wanyi,mandil,lihuaijun,lucida@cmbchina.com}
}
\maketitle              % typeset the header of the contribution
\begin{abstract}
Large Language Models (LLMs) have made significant strides in natural language processing and generation, yet their ability to handle long-context input remains constrained by the quadratic complexity of attention computation and linear-increasing key-value memory footprint. To reduce computational costs and memory, key-value cache compression techniques are commonly applied at inference time, but this often leads to severe performance degradation, as models are not trained to handle compressed context. Although there are more sophisticated compression methods, they are typically unsuitable for post-training because of their incompatibility with gradient-based optimization or high computation overhead. To fill this gap with no additional parameter and little computation overhead, we propose Lag-Relative Sparse Attention(LRSA) anchored by the LagKV compression method for long context post-training. Our method performs chunk-by-chunk prefilling, which selects the top K most relevant key-value pairs in a fixed-size lagging window, allowing the model to focus on salient historical context while maintaining efficiency. Experimental results show that our approach significantly enhances the robustness of the LLM with key-value compression and achieves better fine-tuned results in the question-answer tuning task.

\end{abstract}
\section{Introduction}

Large Language Models (LLMs) have demonstrated impressive capabilities in a wide range of tasks, including dialogue, summarization, code generation, and multistep reasoning~\cite{naveed2023comprehensive,chang2024survey}. As their applications expand, there is a growing need for LLMs to process much longer contexts, such as entire documents, large codebases, and multi-turn conversations, to support more complex and realistic use cases. However, the ability of current models to scale to long-context inputs is fundamentally limited by the quadratic complexity of self-attention with respect to sequence length~\cite{wang2024beyond}. To mitigate these costs, compression techniques such as truncation, token merging, and memory selection are often applied during inference~\cite{wang2024svd,tan2025tokencarve,li2025cmt}. Although these methods can extend the usable context window, they frequently lead to a significant drop in model performance because the underlying models are not trained to handle compressed inputs.

A common workaround for memory bottlenecks in long-context scenarios is to apply key-value (KV) cache compression in inference time~\cite{ge2023model}, selectively retaining only a portion of past tokens. While such heuristics can reduce memory usage and computation burden, they often lead to notable performance degradation, especially in tasks requiring precise recall of earlier content~\cite{li2024survey}. This is largely because models are not trained to operate with compressed contexts, resulting in a mismatch between training and inference behavior. Various long-context compression techniques have been proposed to reduce inference-time memory and computation, such as sparse attention, retrieval-based memory, or low-rank approximations. But these methods are rarely integrated into the post-training process~\cite{lou2024sparser,wang2025recursively,li2023losparse}, either due to their incompatibility with gradient-based optimization or their high computational cost during training. 

To address this train–test mismatch, we propose Lag-Relative Sparse Attention (LRSA), a novel attention mechanism anchored by the LagKV compression method ~\cite{liang2025lagkvlagrelativeinformationkv} designed specifically for long-context post-training. Unlike the prior methods, LSRA introduces a structured sparsity pattern that operates in a chunk-wise manner, where each new input chunk selectively attends to a top-K subset of historical key-value pairs from a fixed-size lagging window. This iterative selection process ensures that only the most relevant past information is considered as the sequences progresses, enabling scalable long-context training. Importantly, our design ensures differentiability and efficiency during training, enabling the model to learn how to operate under this sparsified context directly. 

We implement our algorithm within the Megatron~\cite{shoeybi2019megatron,megatronlm} and MindSpeedLLM framework~\cite{mingspeedllm}, a framework for training large-scale language models. 

% ~\ref{fig1}

% articles~\cite{ref_article1}, an LNCS chapter~\cite{ref_lncs1}, a
% book~\cite{ref_book1}, proceedings without editors~\cite{ref_proc1},
% and a homepage~\cite{ref_url1}. Multiple citations are grouped
% \cite{ref_article1,ref_lncs1,ref_book1},
% \cite{ref_article1,ref_book1,ref_proc1,ref_url1}.

\begin{credits}
% \subsubsection{\ackname} This study was funded by X (grant number Y).
% A bold run-in heading in small font size at the end of the paper is
% used for general acknowledgments, for example: 

\subsubsection{\discintname}
% (optional) acknowledgments\footnote{If EquinOCS, our proceedings submission
% system, is used, then the disclaimer can be provided directly in the system.},
The authors have no competing interests to declare that are
relevant to the content of this article.
% Or: Author A has received research
% grants from Company W. Author B has received a speaker honorarium from
% Company X and owns stock in Company Y. Author C is a member of committee Z.
\end{credits}

\section{Related Works}

\subsection{Token Eviction and Context Compression}
Most large language models (LLMs) are entirely based on the self-attention mechanism~\cite{vaswani2017attention} to assess the relevance of historical tokens to predict the next token. As a result, many KV compression and token eviction strategies are designed around attention-based heuristics, which identify and drop less important tokens~\cite{zhang2023h2o,liu2023scissorhands,li2024snapkv,kvpress}. These approaches often maintain strong performance even at high compression ratios. However, their effectiveness often depends on access to the final query or instruction to compute accurate importance scores~\cite{li2024scbench,feng2024ada,tang2024razorattention}. More importantly, attention-based eviction is fundamentally incompatible with FlashAttention (FA)~\cite{dao2022flashattention}, which precomputes attention in a way that does not allow selective token dropping based on attention scores. This incompatibility makes such methods impractical for real-world deployment, especially in long-context scenarios where FA is essential for efficient inference.

\subsection{Sparse Attention}
The high computational cost of full attention has motivated the development of sparse attention mechanisms that reduce complexity while preserving model quality. Early work such as Sparse Transformer~\cite{child2019generating} introduced fixed attention patterns that skip irrelevant positions, while later methods like BigBird~\cite{zaheer2020big} and Longformer~\cite{beltagy2020longformer} use global + local patterns to maintain scalability. More adaptive approaches, including SnapKV~\cite{li2024snapkv} and AnchorAttention~\cite{zhang2025anchorattention}, dynamically prune KV pairs based on cumulative attention scores. Despite improvements, these strategies often focus on inference time efficiency and are rarely integrated into the pre-training process, limiting their robustness under compressed conditions. StreamingLLM~\cite{xiao2023efficient} and FlexPrefill~\cite{lai2025flexprefill} demonstrate further inference optimizations by discarding intermediate tokens or dynamically selecting patterns, but remain inference-only solutions and lack global selection mechanisms. Meanwhile, Native Sparse Attention (NSA)~\cite{yuan2025native} injects compressed information through a gating mechanism that will introduce additional parameters.

\subsection{Positional Embeddings}
Rotary Position Embeddings (RoPE)~\cite{su2024roformer} have emerged as a powerful alternative to traditional absolute or learned positional embeddings. RoPE encodes position information directly into the attention mechanism via complex rotation, enabling models to generalize better to longer contexts and support extrapolation. Recent works such as YaRN~\cite{peng2023yarn} and LongRoPE~\cite{ding2024longrope} have explored modifications to RoPE for further extension of context windows. Our approach inherits standard RoPE but adapts its scaling and interpolation to maintain stability under long sequences and varying token eviction schedules.

\section{Method}
In this section, we present LRSA, Lag-Relative Sparse Attention, which extends LagKV compression from model inference to model training. 
\subsection{Preliminaries}
LLMs' next token prediction relies on the previous tokens. First, in the prefill stage, the model uses its tokenizer to convert the words to $n$ indices of the embedding metrics $E\in \mathbb{R}^{V \times d}$ of the model and collects the representations to form a input matrix, $X\in \mathbb{R}^{n \times d}$. This matrix is the initial tokens of the first layer of LLM and then each layer will output a same shape matrix as next layer's input.
To depict the operations in each layer, we follow the notation system from ~\cite{liu2023dejavucontextualsparsity} with $h$ attention heads.
For each head $i \in [1, h]$ and head dimension $d_h$, we focus on the Query, Key, and Value states, which are converted from tokens by three linear transformation matrices $W_i^Q$, $W_i^K$, $W_i^V \in \mathbb{R}^{d \times d_h}$ separately:
\begin{equation}
	Q_i=XW_i^Q,K_i = XW_i^K,V_i=XW_i^V
\end{equation}
The output $Y \in \mathbb{R}^{n \times d}$ is computed using the attention weights $A_i \in \mathbb{R}^{n \times n}$ and the final output matrix $W^O \, \in \mathbb{R}^{d \times d}$:
\begin{equation}
	Y = Concat_{i \in [1, h]}( A_i V_i )W^O,
\end{equation}
\begin{equation}
\label{eq:attnmask}
A_i =\text{softmax}(\frac{Q_{i}K_i^{T}}{\sqrt{d_h}} + M_i).
\end{equation}
where $A_i$ is the attention weight and $M_i$ is the attention masked matrix. The masked elements of $M_i$ will be filled by negative infinite to eliminate the impaction of unselected tokens.
In LLM where predicting tokens on preceding tokens, the masked matrix is a triangular matrix with the upper triangle masked out.

\subsection{LRSA}
In the work of LagKV~\cite{liang2025lagkvlagrelativeinformationkv}, it proposes to use the lag-relative information to determine the importance of historical tokens. The LagKV token importance score is relying on the token-wise min-max from the subsequent KV chunk to normalize the current one as the following equations:
\begin{equation}
min_i^{p,Z} = min_{seq}({Z_i^{p+1}})
\label{eq:min}
\end{equation}
\begin{equation}
max_i^{p,Z}=max_{seq}({Z_i^{p+1}})
\label{eq:max}
\end{equation}
\begin{equation}
\bar{Z_i^p}=\frac{{Z_i^p}-min_i^{p,Z}}{max_i^{p,Z}-min_i^{p,Z}}
\end{equation}
\begin{equation}
score(Z_i)=Softmax(Std.(\bar{Z_i}))
\end{equation}
\begin{equation}
score_{i}=score(K_i)+score(V_i)
\label{eq:scoresum}
\end{equation}
where the KV sequence is divided into sink part with size $S$ and chunks with lag size $L$, $Z$ is one of $\{K, V\}$, $p$ denotes the chunk index, $i$ represents the attention head index and $seq$ for the sequence axis. Based on $score_i$ and the retention ratio $r$, select the top $rL$ tokens to remain in the KV cache to complete the compression process. 

\begin{figure*}[t]
    \centering
    \includegraphics[width=1\linewidth]{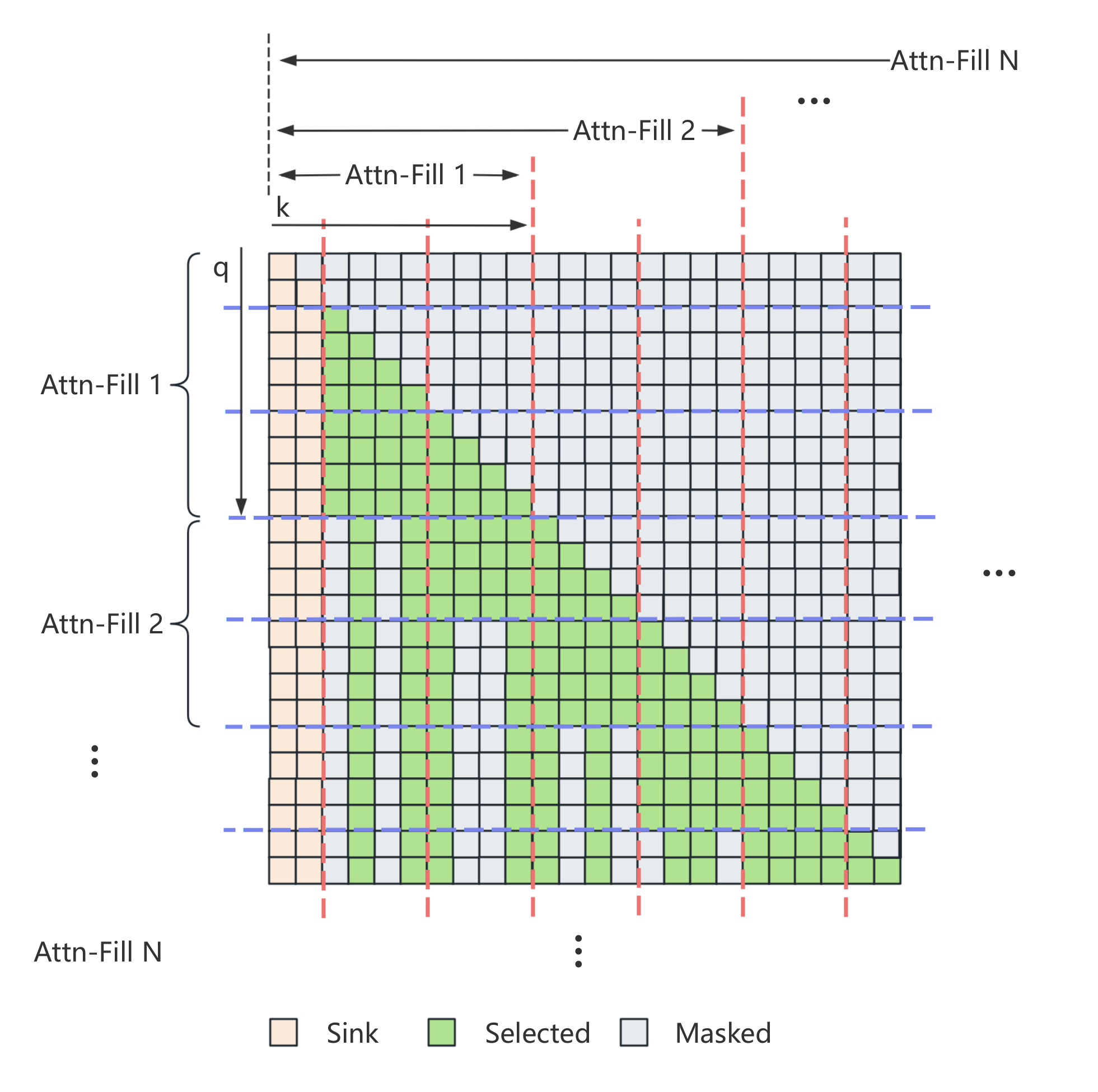}
    \caption{Lag-Relative Sparse Attention: Chunked-Mask is generated by the LagKV algorithm.}
    \label{fig:lrsa}
\end{figure*}

The LagKV work also demonstrates that prefilling long contexts in a chunk-by-chunk manner not only preserves the model's core capabilities but also enhances efficiency by reducing the number of tokens involved in attention computation. Following their approach, the LRSA mask matrix takes the form like Figure~\ref{fig:lrsa}.

For efficiency, the prefilling process processes multiple chunks simultaneously. The number of filling chunks can be optimized up to different hardware and context lengths. For instance, with two chunks at a time:
\begin{itemize}
\item {\bf Attention Filling 1 (Attn-Fill 1)}: The sink portion and two lag-sized KV segments are fed into the attention calculation. The attention mask is a full-size matrix with the upper triangle masked out. Meanwhile, the KV cache is updated with the first chunk compressed and the second chunk retained as a sliding window.

\item {\bf Attention Filling 2 (Attn-Fill 2)}: The next two chunks are concatenated to the compressed KV cache. The third chunk's queries require no additional masking, while the fourth chunk's queries are masked for the second chunk's portion. The masked indices are provided by the new compression process.

\item {\bf Repeat} step 2 until all chunks processed.
\end{itemize}

Our masking strategy differs fundamentally from approaches like NSA~\cite{yuan2025native} or Quest~\cite{10.5555/3692070.3694025}. Their masking is query-dependent, requiring all key-value (KV) pairs to remain in the cache since different queries may attend to different KVs. In contrast, LRSA employs a static mask for subsequent query chunks because the LagKV scoring method is entirely query-independent. This masking strategy allows the KV cache to be condensed shorter.

\section{Experiment}

\subsection{Setups}
\textbf{Models} Our evaluation is conducted on Qwen2.5-1.5B-Base~\cite{qwen2025qwen25technicalreport}, which support up to 32K context length in their pre-trained form. We applied the same evaluation framework to the pre-trained form without fine-tuning and the fine-tuned form to make a fairly comparison. 

\textbf{Benchmark} We evaluate the models with RULER~\cite{hsieh2024ruler}, which generates synthetic examples to evaluate long-context language models with configurable sequence length and task complexity. 

\textbf{Datasets} We create the pre-training datasets using four datasets open-sourced by LongAlign and LongLora. Here are the basic information of the four basic datasets:
\begin{enumerate}
    \item THUDM/LongAlign-10k This dataset includes 10,000 long-text tasks, with approximately 10\% of the samples written in Chinese.
    \item Yukang/LongAlpaca-12k LongAlpaca comprises 12,000 long-text tasks, LongAlpaca primarily focuses on reading comprehension of academic papers, with a mix of shorter samples included to maintain dataset balance.
    \item wenbopan/RefGPT-Fact-v2-8x is an expanded version of the Mutonix/RefGPT-Fact-v2 dataset, which features high-quality conversations centered on document extraction and understanding. Since the original dataset was relatively short, additional longer samples were synthesized to enhance its length.
    \item wenbopan/anti-haystack is a set of long-text tasks generated using GPT-4, emphasizing symbolic reasoning, accurate fact recall, and precise paragraph referencing.
\end{enumerate} 

The datasets are structured in a question–answer format. Since not all data samples naturally reach the target length of 32K tokens, we perform synthetic augmentation to extend their lengths. Specifically, we categorize samples into length-based groups. For samples that fall short of 32K tokens, we randomly select supplementary paragraphs from longer samples within the same or compatible groups. To preserve semantic coherence and ensure the consistency of the question content, we prepend a new document section and use the delimiter "Document" to clearly separate concatenated texts.

\textbf{Baselines}. We evaluate two baseline models to assess the effectiveness of our proposed training strategy:
(i) \textbf{Qwen2.5-1.5B-base}, the original pretrained model without any additional fine-tuning;
(ii) Fine-tuned \textbf{Qwen2.5-1.5B-base}, which is trained on our curated dataset without applying any compression algorithm.

\textbf{Implementation} All experiments are conducted on Ascend 910B NPU with 64GB memory, leveraging parallelisms implemented in the Megatron distributed training framework. We use the following fine-tuning hyperparameters:
\begin{itemize}
    \item 4000 iterations with global batch size of 16.
    \item Train in bfloat16 precision with a learning rate of 1.25e-5 without the warm-up stage and then decayed to 0 throughout training following a cosine schedule.
    \item Use the AdamW optimizer~\cite{loshchilov2017decoupled} with $\beta1 = 0.9$, $\beta2 = 0.95$ and a weight decay of $1e-1$.
    \item For the configuration of LagKV compression, delay size $L=1024$ and retention ratio $r=0.5$.
\end{itemize} 

\begin{figure*}[h]
    \centering
    \includegraphics[width=1\linewidth]{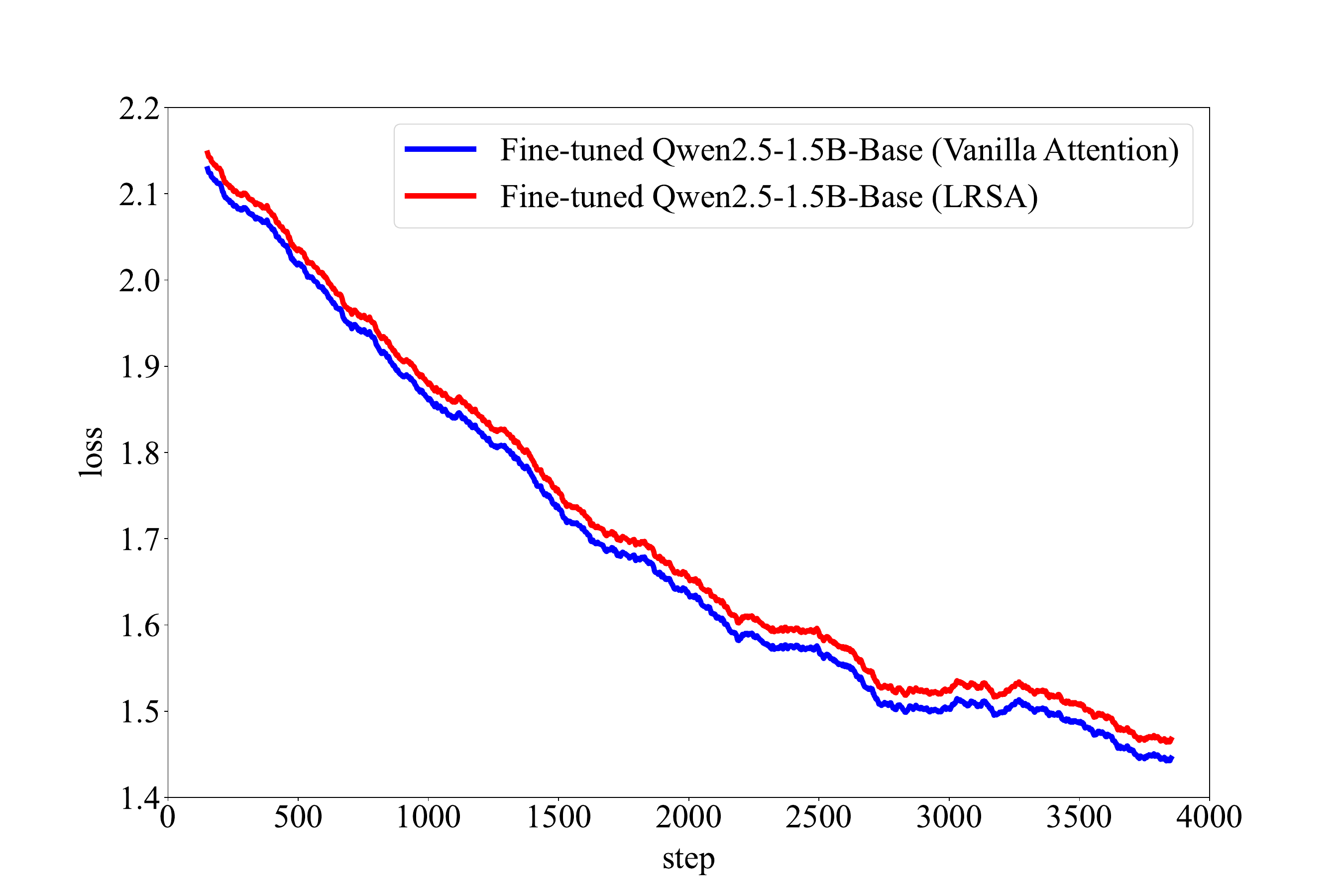}
    \caption{Training Loss of Fine-tuned Qwen2.5-1.5B-Base (w/o LRSA)}
    \label{fig:loss}
\end{figure*}

\subsection{Training Loss}
Fig~\ref{fig:loss} compares the training loss curves of two fine-tuned Qwen2.5-1.5B-Base models: one using standard full attention (blue) and the other employing our proposed Lag-Relative Sparse Attention (LRSA, red). Both models are trained on the same dataset and under identical training conditions. Although the LRSA-enhanced model shows a slightly higher loss throughout the training process, the gap remains small and stable. This shows that our sparse attention mechanism can be integrated into training without causing significant degradation in convergence, thus enabling efficient long-context modeling while maintaining training stability.

\subsection{RULER Benchmark}
The main results on the RULER benchmark are presented in Table~\ref{tab:main_table}.

\begin{table*}[ht!]
\centering
\caption{Results of RULER benchmark.}

\resizebox{\textwidth}{!}{
\begin{tabular}{c|c|ccccccccccccc}
\toprule
\textbf{Model}      & \textbf{Method} & NS1 & NS2 & NS3 & NMK1 & NMK2 & NMK3 & NMV & NMQ & VT & CWE & FWE & QA1 & QA2 \\
\midrule
\multirow{12}{*}{\rotatebox[origin=c]{90}{\parbox{3cm}{\centering Qwen2.5-1.5B-Base}}}
& $L_S$=4k,r=1x,L=128 & 100.0 & 100.0 & 100.0 & 99.4 & 97 & 81.2 & 96.9 & 62.3 & 92.4 & 68.4 & 75.73 & 71.8 & 42.4\\
& $L_S$=4k,r=2x,L=128 & 100.0 & 94.0 & 25.0 & 85.2 & 44.4 & 3.6 & 87.5 & 56.6 & 91.2 & 40.0 & 71.6 & 63.2 & 36.2\\
& $L_S$=4k,r=4x,L=128 & 98.6 & 64.0 & 4.2 & 62.6 & 16.4 & 0.2 & 54.1 & 39.1 & 90.6 & 15.5 & 59.5 & 56.6 & 31.4\\
\cline{2-15}
& $L_S$=8k,r=1x,L=256 & 100.0 & 100.0 & 100.0 & 97.6 & 92.0 & 69.6 & 93.6 & 79.1 & 87.4 & 25.8 & 65.6 & 52.8 & 38\\
& $L_S$=8k,r=2x,L=256 & 100.0 & 96.8 & 52.8 & 87.8 & 35.0 & 5.6 & 86.9 & 65.7 & 86.6 & 18.6 & 59.7 & 45.6 & 36.2\\
& $L_S$=8k,r=4x,L=256 & 99.2 & 66.2 & 10.0 & 64.6 & 15.0 & 1.2 & 55.7 & 43.7 & 90.3 & 10.4 & 49.8 & 36.8 & 28.4\\
\cline{2-15}
& $L_S$=16k,r=1x,L=512 & 100.0 & 100.0 & 99.8 & 98.0 & 83.2 & 69.0 & 89.9 & 87.6 & 83.6 & 20.6 & 74.1 & 49.6 & 37.2\\
& $L_S$=16k,r=2x,L=512 & 100.0 & 91.4 & 44.4 & 82.8 & 23.0 & 8.4 & 82.8 & 73.0 & 88.5 & 7.1 & 70.7 & 43.0 & 33.4\\
& $L_S$=16k,r=4x,L=512 & 99.8 & 73.8 & 14.0 & 60.0 & 7.2 & 1.6 & 62.8 & 52.0 & 87.4 & 3.4 & 65.9 & 33.6 & 31.2\\
\cline{2-15}
& $L_S$=32k,r=1x,L=1024 & 100.0 & 98.8 & 99.4 & 92.2 & 76.6 & 44.2 & 80.9 & 74.0 & 83.3 & 3.7 & 65.1 & 46.2 & 36.0\\
& $L_S$=32k,r=2x,L=1024 & 100.0 & 92.8 & 41.0 & 78.8 & 20.2 & 6.0 & 69.6 & 68.6 & 87.0 & 3.2 & 67.7 & 34.8 & 31.2\\
& $L_S$=32k,r=4x,L=1024 & 98.4 & 72.2 & 10.8 & 59.0 & 5.0 & 1.8 & 47.1 & 44.4 & 81.7 & 1.0 & 62.9 & 28.2 & 28.8\\
\midrule
\multirow{12}{*}{\rotatebox[origin=c]{90}{\parbox{3cm}{\centering Qwen2.5-1.5B-Base\\(Fine-tuned by Vanilla Attention)}}}
& $L_S$=4k,r=1x,L=128 & 100.0 & 99.8 & 99.8 & 98.6 & 94.0 & 81.6 & 92.5 & 62.3 & 93.8 & 47.6 & 73.3 & 76.4 & 44.8\\
& $L_S$=4k,r=2x,L=128 & 99.4 & 92.8 & 25.0 & 82.6 & 43.0 & 6.0 & 78.4 & 55.9 & 92.5 & 34.8 & 67.5 & 66.2 & 40.2\\
& $L_S$=4k,r=4x,L=128 & 99.0 & 62.6 & 4.8 & 49.6 & 12.2 & 0.2 & 37.9 & 27.7 & 90.0 & 18.1 & 55.6 & 53.2 & 33.8\\
\cline{2-15}
& $L_S$=8k,r=1x,L=256 & 100.0 & 99.8 & 99.8 & 94.6 & 86.8 & 65.6 & 85.6 & 67.3 & 62.4 & 22.3 & 62.8 & 57.6 & 41.8\\
& $L_S$=8k,r=2x,L=256 & 100.0 & 95.0 & 45.4 & 78.6 & 32.0 & 6.6 & 76.6 & 59.2 & 68.8 & 21.8 & 55.2 & 48.6 & 34.6\\
& $L_S$=8k,r=4x,L=256 & 99.6 & 70.6 & 10.8 & 47.4 & 12.6 & 2.6 & 43.2 & 32.8 & 80.8 & 12.1 & 46.4 & 36.4 & 31.0\\
\cline{2-15}
& $L_S$=16k,r=1x,L=512 & 100.0 & 99.6 & 99.4 & 88.6 & 72.8 & 50.6 & 83.2 & 62.4 & 61.2 & 8.5 & 70.2 & 57.6 & 38.0\\
& $L_S$=16k,r=2x,L=512 & 99.6 & 91.8 & 38.4 & 67.0 & 21.0 & 7.8 & 73.6 & 57.0 & 79.8 & 9.2 & 69.8 & 46.8 & 34.0\\
& $L_S$=16k,r=4x,L=512 & 99.6 & 69.2 & 8.2 & 48.4 & 6.8 & 4.6 & 47.7 & 37.4 & 74.6 & 5.2 & 60.0 & 36.4 & 28.4\\
\cline{2-15}
& $L_S$=32k,r=1x,L=1024 & 100.0 & 99.4 & 98.2 & 85.4 & 57.6 & 30.6 & 72.9 & 54.8 & 52.9 & 8.2 & 64.6 & 42.4 & 34.4\\
& $L_S$=32k,r=2x,L=1024 & 100.0 & 90.4 & 31.6 & 73.0 & 15.4 & 5.4 & 55.0 & 44.0 & 65.2 & 7.7 & 67.3 & 34.8 & 31.6\\
& $L_S$=32k,r=4x,L=1024 & 99.8 & 65.8 & 6.8 & 44.2 & 6.6 & 3.4 & 35.1 & 27.5 & 72.8 & 5.8 & 60.1 & 30.0 & 28.2\\
\midrule
\multirow{12}{*}{\rotatebox[origin=c]{90}{\parbox{3cm}{\centering Qwen2.5-1.5B-Base\\(Fine-tuned by LRSA)}}}
& $L_S$=4k,r=1x,L=128 & 100.0 & 99.8 & 100.0 & 97.6 & 96.8 & 88.8 & 44.2 & 63.6 & 95.7 & 49.3 & 73.9 & 74.0 & 45.0\\
& $L_S$=4k,r=2x,L=128 & 100.0 & 96.0 & 53.0 & 91.2 & 62.4 & 17.6 & 85.7 & 59.6 & 90.8 & 39.1 & 71.5 & 66.4 & 43.0\\
& $L_S$=4k,r=4x,L=128 & 99.2 & 80.6 & 4.4 & 68.0 & 27.4 & 1.0 & 53.4 & 41.9 & 92.5 & 23.4 & 60.7 & 56.4 & 35.6\\
\cline{2-15}
& $L_S$=8k,r=1x,L=256 & 100.0 & 100.0 & 99.6 & 94.2 & 89.0 & 79.8 & 79.6 & 63.5 & 62.4 & 21.2 & 61.3 & 54.6 & 42.0\\
& $L_S$=8k,r=2x,L=256 & 100.0 & 99.0 & 65.4 & 87.6 & 53.2 & 17.2 & 77.9 & 60.7 & 61.7 & 22.7 & 58.4 & 53.6 & 37.0\\
& $L_S$=8k,r=4x,L=256 & 99.8 & 86.6 & 20.4 & 67.6 & 23.0 & 1.6 & 58.5 & 42.0 & 88.8 & 14.8 & 51.2 & 39.8 & 31.6\\
\cline{2-15}
& $L_S$=16k,r=1x,L=512 & 100.0 & 99.4 & 99.6 & 85.4 & 70.6 & 64.0 & 76.3 & 63.7 & 57.2 & 4.4 & 73.8 & 56.4 & 40.2\\
& $L_S$=16k,r=2x,L=512 & 100.0 & 98.6 & 56.8 & 77.6 & 35.4 & 17.2 & 74.8 & 59.5 & 70.24 & 4.7 & 72.1 & 50.4 & 38.8\\
& $L_S$=16k,r=4x,L=512 & 100.0 & 80.6 & 18.0 & 55.2 & 13.2 & 5.6 & 58.7 & 43.9 & 76.4 & 3.1 & 69.8 & 40.4 & 31.2\\
\cline{2-15}
& $L_S$=32k,r=1x,L=1024 & 100.0 & 94.0 & 95.6 & 78.6 & 54.8 & 56.4 & 63.1 & 49.3 & 50.1 & 3.0 & 70.0 & 46.2 & 37.2\\
& $L_S$=32k,r=2x,L=1024 & 100.0 & 94.0 & 47.2 & 75.2 & 21.6 & 14.2 & 56.6 & 46.0 & 58.8 & 3.0 & 70.8 & 41.0 & 34.2\\
& $L_S$=32k,r=4x,L=1024 & 100.0 & 80.4 & 18.0 & 56.6 & 7.8 & 3.6 & 45.2 & 33.7 & 66.8 & 3.1 & 70.1 & 31.6 & 27.4\\
\bottomrule
% \vspace{-1em}

\end{tabular}
}
\label{tab:main_table}
\footnotesize\textit{Note: $L_S$ represents the context length, r represents the retention ratio, and L represents the lag size.} 
\end{table*}

\begin{figure*}[h!]
    \centering
    \includegraphics[width=0.9\linewidth]{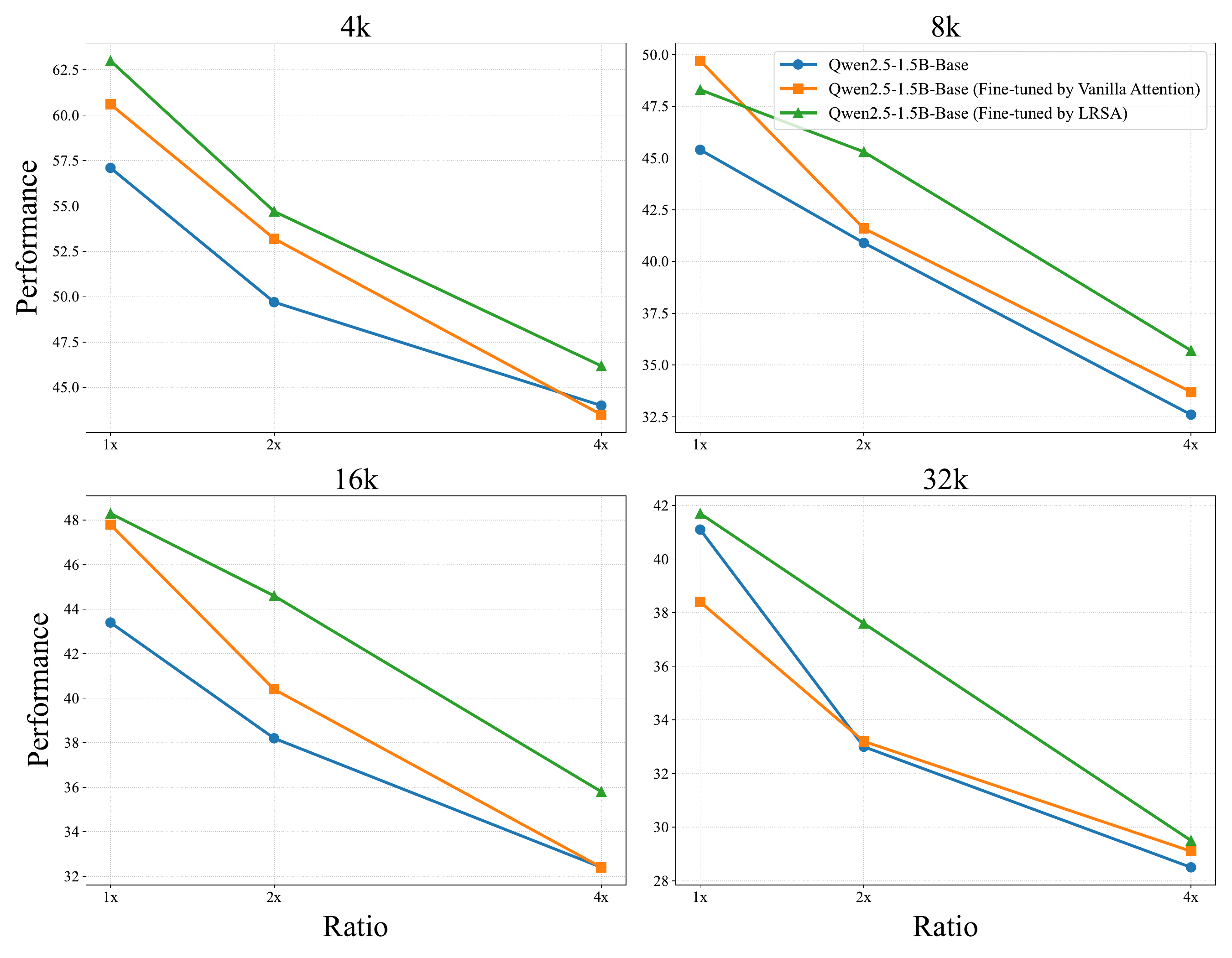}
    \caption{Question-Answer-focused Statistics from RULER benchmark}
    \label{fig:qa}
\end{figure*}

\begin{figure*}[h]
    \centering
    \includegraphics[width=0.9\linewidth]{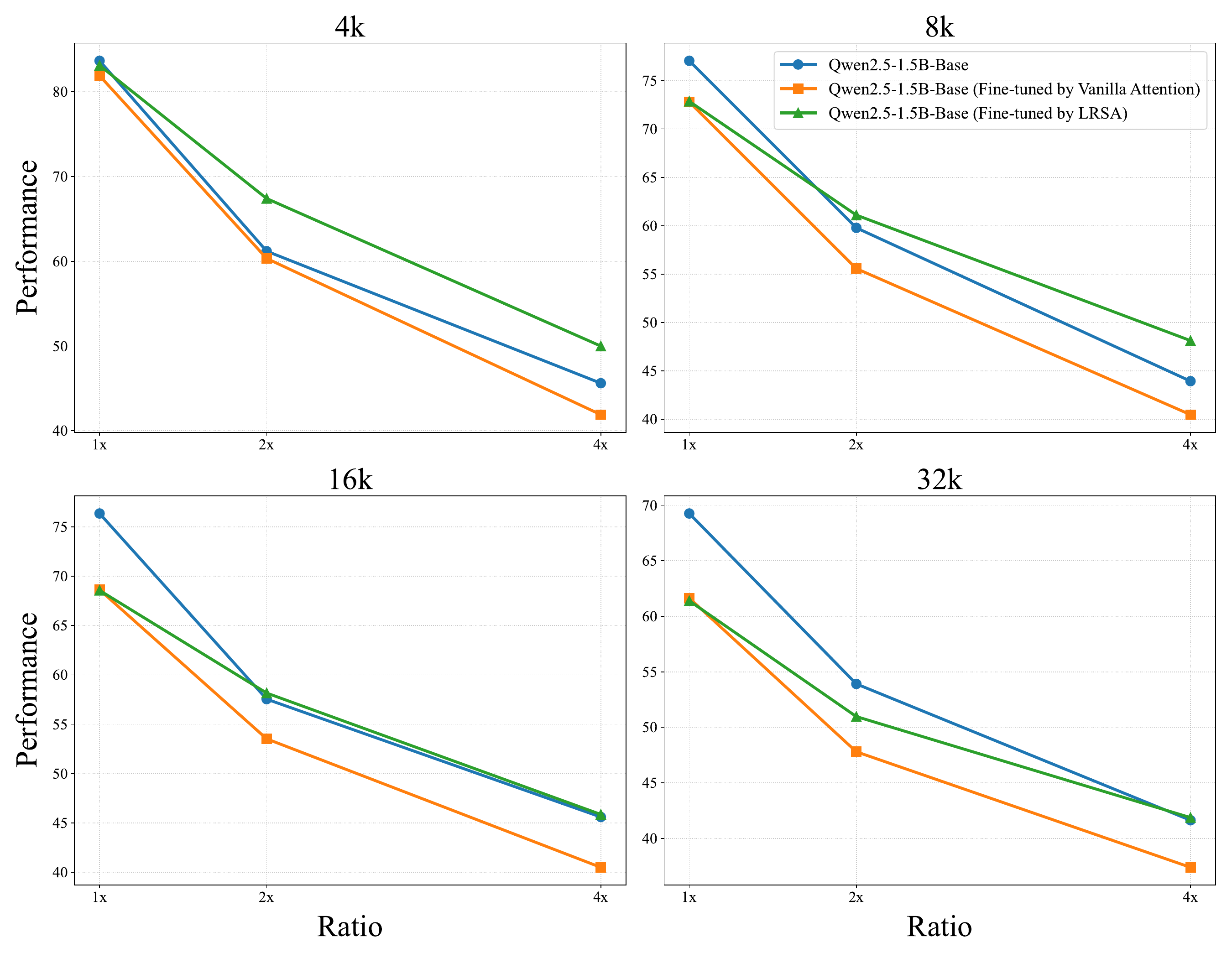}
    \caption{Average Statistics of RULER benchmark}
    \label{fig:average}
\end{figure*}
Fig.\ref{fig:qa} and Fig.\ref{fig:average} show the performance of different Qwen2.5-1.5B variants on question-answer (QA) tasks and overall synthetic tasks, respectively, under varying compression ratios and input sequence lengths. Several key insights can be drawn from these results:
\begin{itemize}
    \item In the specific QA training task, Fig.\ref{fig:qa} shows that LRSA consistently outperforms Vanilla attention with or without compression. It demonstrates that LRSA can effectively achieve better training results with fewer tokens attendance, even the loss is slightly higher.
    \item  LRSA improves the robustness of compression. The fine-tuned model with LRSA shows a considerably slower performance degradation as the compression ratio times increase from $1\times$ to $4\times$ in RULER and specific QA tasks. For example, at 8k and 16k input sequence lengths, it maintains a clear margin above the others at both $2\times$ and $4\times$ compression. This indicates that LRSA helps the model to better adapt to compressed-memory scenarios by teaching it to prioritize and retain salient tokens during training. 
    \item Quality of our QA-aligned fine-tuning dataset. Both fine-tuned models surpass the base model, reflecting the strength of the QA-style instruction dataset. 
\end{itemize}

\section{Conclusion}

In this work, we introduce Lag-Relative Sparse Attention (LRSA), a lightweight, plug-in-go, and effective post-training mechanism tailored for long-context language modeling under memory-constrained conditions. Built upon the LagKV compression strategy, LRSA enables LLMs to selectively retain and attend to the most salient key-value pairs within a fixed size window, thereby reducing the memory and compute overhead of full attention while maintaining differentiability and training efficiency. The results of our experiments reveal that the fine-tuned model with LRSA outperforms the one with vanilla attention to QA-related evaluation sets. Moreover, the average score of the fine-tuned model with LRSA the RULER synthetic evaluation set is higher than that with vanilla attention. In addition, we show that integrating LRSA into the Megatron framework is feasible and stable, with only minimal impact on convergence behavior. These findings suggest that LRSA is a practical and generalizable solution for large scale models on the post-training stage to handle long-context input.
\bibliographystyle{splncs04}
\bibliography{mybibliography}




% LNCS Homepage, \url{http://www.springer.com/lncs}, last accessed 2023/10/25
% \end{thebibliography}
\end{document}